\def\year{2020}\relax
\documentclass[letterpaper]{article} % DO NOT CHANGE THIS
\usepackage{aaai20}  % DO NOT CHANGE THIS
\usepackage{times}  % DO NOT CHANGE THIS
\usepackage{helvet} % DO NOT CHANGE THIS
\usepackage{courier}  % DO NOT CHANGE THIS
\usepackage[hyphens]{url}  % DO NOT CHANGE THIS
\usepackage{graphicx} % DO NOT CHANGE THIS
\usepackage{amsmath}
\usepackage{algorithm}
\usepackage[noend]{algpseudocode}
\usepackage{graphicx}  % DO NOT CHANGE THIS
\newcommand{\commentout}[1]{}
\title{ALOHA: Artificial Learning of Human Attributes for Dialogue Agents}
\renewcommand*{\thefootnote}{\fnsymbol{footnote}}

\author{Aaron W. Li,\textsuperscript{\rm 1}
	Veronica Jiang\footnotemark[1],\textsuperscript{\rm 1}
	Steven Y. Feng\footnotemark[1],\textsuperscript{\rm 1}
	Julia Sprague,\textsuperscript{\rm 1}
	Wei Zhou,\textsuperscript{\rm 2}
	Jesse Hoey\textsuperscript{\rm 1}
	\\
	\textsuperscript{\rm 1}David R. Cheriton School of Computer Science\\ University of Waterloo\\ Waterloo, Ontario, Canada\\
	\textsuperscript{\rm 2}Huawei Technologies Co., Ltd.\\
	\{w89li, r4jiang, sy2feng, jsprague\}@uwaterloo.ca, wei.zhou1@huawei.com, jhoey@uwaterloo.ca
}

\begin{document}
	\maketitle
	
	\begin{abstract}
		For conversational AI and virtual assistants to communicate with humans in a realistic way, they must exhibit human characteristics such as expression of emotion and personality. Current attempts toward constructing human-like dialogue agents have presented significant difficulties. We propose Human Level Attributes (HLAs) based on {\em tropes} as the basis of a method for learning dialogue agents that can imitate the personalities of fictional characters. Tropes are characteristics of fictional personalities that are observed recurrently and determined by viewers' impressions. By combining detailed HLA data with dialogue data for specific characters, we present a dataset, HLA-Chat, that models character profiles and gives dialogue agents the ability to learn characters' language styles through their HLAs. We then introduce a three-component system, ALOHA (which stands for Artificial Learning of Human Attributes), that combines character space mapping, character community detection, and language style retrieval to build a character (or personality) specific language model. Our preliminary experiments demonstrate that two variations of ALOHA, combined with our proposed dataset, can outperform baseline models at identifying the correct dialogue responses of chosen target characters, and are stable regardless of the character's identity, the genre of the show, and the context of the dialogue.
	\end{abstract}
	
	\footnotetext[1]{Authors contributed equally}
	\renewcommand*{\thefootnote}{\arabic{footnote}}
	
	\section{Introduction}
	%The success and growth of conversational AI and virtual assistants rest in their ability to connect with humans interactively. Diction, tone, and speech patterns are key elements needed for them to communicate more effectively, which many systems lack today. Given the trajectory of the current models available, there is an inevitable plateau unless they are able to better imitate human-like qualities. There has been significant research on the generation of dialogue for chatbots, but lack of research done on how to emulate these human-like characteristics is limiting their growth.
	
	%this seems off topic for the paper
	%A specific use-case of conversational AI is in the development of virtual assistants with adjustable socio-emotional personalities to aid the ongoing effort to construct assistive technologies for cognitively disabled patients. In these situations, subtly adjusting the emotional delivery of text can have strong effect on the adoption of the technologies (Robillard et al.,2018) [CITE]. It has been shown that attention increases when there is socio-emotional engagement in conversation [CITE  Socio-emotional  engagement,  joint  attention,  imitation,  and  conversation  skill:  Analysis  in  typical development and specific language impairment], leading to more consistent conversations and stronger engagement.
	%%SF: Agreed, let's keep this part out
	
	Attempts toward constructing human-like dialogue agents have met significant difficulties, such as maintaining conversation consistency~\cite{zhang2018personalizing}. This is largely due to inabilities of dialogue agents to engage the user emotionally because of an inconsistent personality~\cite{rashkin2019towards}. Many agents use personality models that attempt to map personality attributes into lower dimensional spaces (e.g. the \textit{Big Five} \cite{john1999big}). However, these represent human personality at a very high-level and lack depth. They prohibit the ability to link specific and detailed personality traits to characters, and to construct large datasets where dialogue is traceable back to these traits.
	
\begin{figure}
		\begin{tabular}{ccc}
			\multicolumn{1}{c}{\includegraphics[width=.95\columnwidth]{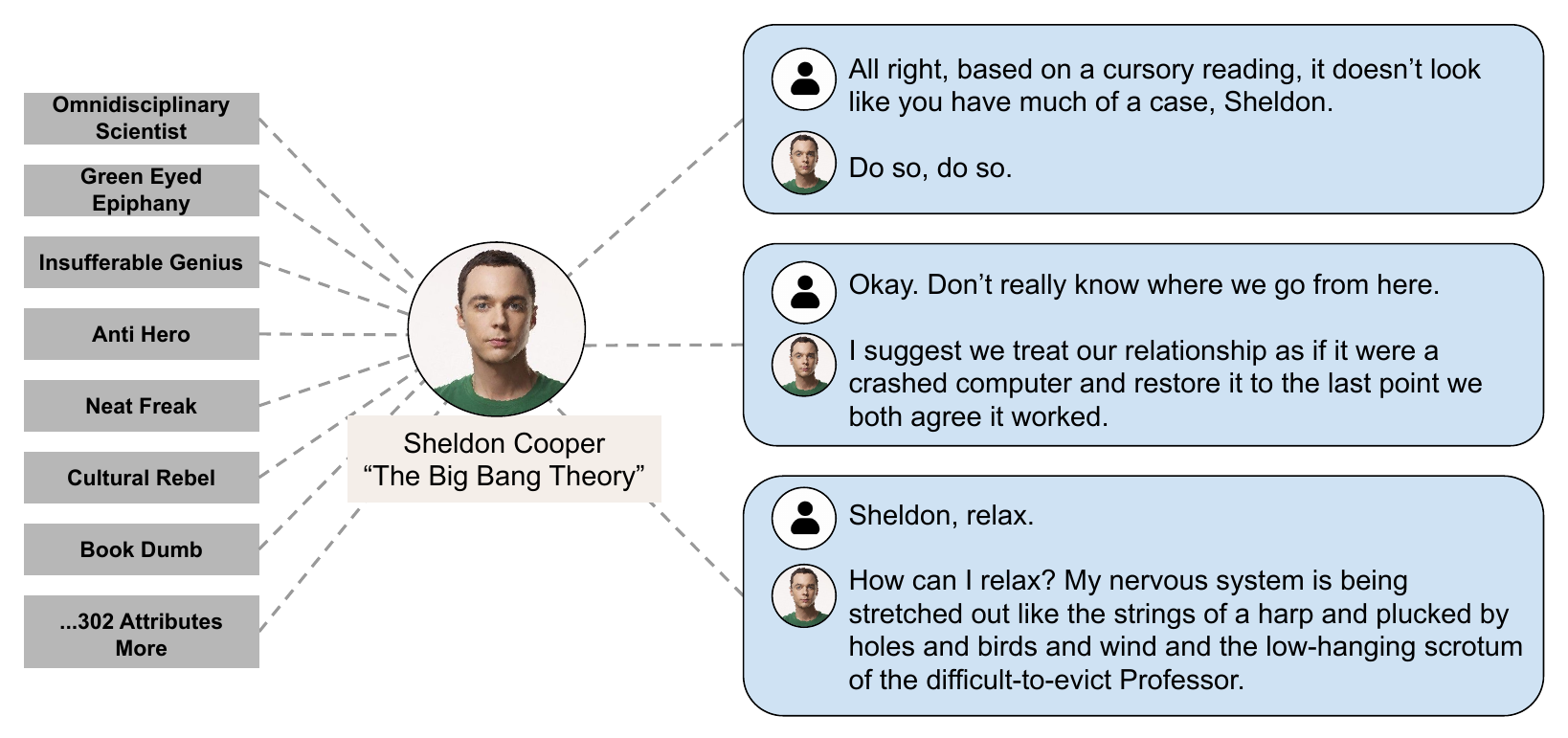}}\\
		\end{tabular}
		\caption{\label{fig:sheldon_diagram} Example of a character and its associated HLAs (tropes) on the left and dialogue lines on the right.}
	\end{figure}

	For this reason, we propose Human Level Attributes (HLAs), which we define as characteristics of fictional characters representative of their profile and identity. We base HLAs on {\em tropes} collected from TV Tropes~\cite{tvtropes}, which are determined by viewers' impressions of the characters. See Figure~\ref{fig:sheldon_diagram} for an example. Based on the hypothesis that profile and identity contribute effectively to language style \cite{pennebaker1999linguistic}, we propose that modeling conversation with HLAs is a means for constructing a dialogue agent with stable human-like characteristics. By collecting dialogues from a variety of characters along with this HLA information, we present a dataset, HLA-Chat, with novel labelling of this dialogue data traceable back to both its context and associated human-like qualities.
	%%JH say where the dialogues come from for this dataset - its not really a novel dataset, but rather a novel labelling of an existing dataset, right? try to clarify that better 
	%%SF: ADDRESSED - please check
	
	We also propose a system called ALOHA (Artificial Learning of Human Attributes) as a novel method of incorporating HLAs into dialogue agents. ALOHA maps characters to a latent space based on their HLAs, determines which are most similar in profile and identity, and recovers language styles of specific characters. We test the performance of ALOHA in character language style recovery against four baselines, demonstrating outperformance and system stability. We also run a human evaluation supporting our results. %We analyze the results in detail.
	
	Our major contributions are: (1) We propose HLAs as personality aspects of fictional characters from the audience's perspective based on {\em tropes}; (2) We provide a large dialogue dataset, HLA-Chat, traceable back to both its context and associated human-like attributes; (3) We propose a system called ALOHA that is able to recommend responses linked to specific characters. We demonstrate that ALOHA, combined with the proposed dataset, HLA-Chat, outperforms baselines. ALOHA also shows stable performance regardless of the character's identity, genre of the show, and context of the dialogue. We release all of ALOHA's data and code along with additional information for reproduction.\footnote{\url{https://github.com/newpro/aloha-chatbot}}
	
	\section{Related Work\label{sec:related_works}}
	%JH2 remove subsections to save space - its very repetitive as you start each section with the same words
	%%SF2: okay
	%\subsection{Task Completion Chatbots (TCC)}
	Task completion chatbots (TCC), or task-oriented chatbots, are dialogue agents used to fulfill specific purposes, such as helping customers book airline tickets, or a government inquiry system. Examples include the AIML based chatbot~\cite{satu2015review} and DIVA Framework \cite{xuetao2009impact}. While TCC are low cost, easily configurable, and readily available, they are restricted to working well for particular domains and tasks.
	
	Open-domain chatbots are more generic dialogue systems. An example is the \textit{Poly-encoder} from \citeauthor{humeau2019real} \shortcite{humeau2019real}. It outperforms the Bi-encoder \cite{mazare2018training,dinan2018wizard} and matches the performance of the Cross-encoder \cite{wolf2019transfertransfo} while maintaining reasonable computation time. It performs strongly on downstream language understanding tasks involving pairwise comparisons, and demonstrates state-of-the-art results on the ConvAI2 challenge \cite{dinan2019second}. \textit{Feed Yourself} \cite{hancock2019learning} is an open-domain dialogue agent with a self-feeding model. When the conversation goes well, the dialogue becomes part of the training data, and when the conversation does not, the agent asks for feedback. Lastly, \textit{Kvmemnn} \cite{eric2017key} is a key-value memory network with a knowledge base that uses a key-value retrieval mechanism to train over multiple domains simultaneously. We use all three of these models as baselines for comparison. While these can handle a greater variety of tasks, they do not respond with text that aligns with particular human-like characteristics.
	
	%\subsection{Human-like Dialogue Agents}
	%With recent advancements in deep learning, modeling conversations are trainable with noisy, open domain data [A Neural Conversational Model]. In addition, the research pointed out the lack of a coherent personality makes it difficult for systems to pass the Turing test [Computing machinery and intelligence]. 
	\citeauthor{li2016persona} \shortcite{li2016persona} defines persona (composite of elements of identity) as a possible solution at the word level, using backpropagation to align responses via word embeddings. \citeauthor{bartl2017retrieval} \shortcite{bartl2017retrieval} uses sentence embeddings and a retrieval model to achieve higher accuracy on dialogue context. \citeauthor{liu2019emotion} \shortcite{liu2019emotion} applies emotion states of sentences as encodings to select appropriate responses. \citeauthor{pichl2018alquist} \shortcite{pichl2018alquist} uses knowledge aggregation and hierarchy of sub-dialogues for high user engagement. \citeauthor{personage} \shortcite{personage}'s PERSONAGE focuses on generating language using the \textit{extraversion} personality trait of the Big Five. However, these agents all represent personality at a high-level and lack detailed human qualities. We model language styles through HLAs which are much more detailed and specific. Hence, the language styles we are recovering may likely capture additional information. %At location, object, and setting level, Jack Urbanek [LIGHT dialogue agent] utilized these ingredients to allow better predictions of agent behaviour and dialogue.
	
	%\subsection{Language Modelling in Fictional Environments}
	%LIGHT \cite{urbanek2019learning} models adventure game characters' dialogues, actions, and emotions. It focuses on the agent identities (e.g. \textit{thief}, \textit{king}, \textit{servant}) which includes limited information on realistic human behaviours. \citeauthor{pasunuru2018game} \shortcite{pasunuru2018game} models online soccer games as dynamic visual context. \citeauthor{wang2016learning} \shortcite{wang2016learning} models user dialogue to complete tasks involving certain configurations of blocks. \citeauthor{antol2015vqa} \shortcite{antol2015vqa} models open-ended questions, but is limited to visual contexts. \citeauthor{bordes2016learning} \shortcite{bordes2016learning} tracks user dialogues but is goal-oriented. \citeauthor{ilinykh2019meetup} \shortcite{ilinykh2019meetup} tracks players' dialogues and movements in a visual environment, and is grounded on navigation tasks. All of these perform well in their respective fictional environments, but are not a strong representation of human dialogue in reality.
	
	\section{Methodology}
	\subsection{Human Level Attributes (HLA)\label{sec:HLA}}
	%We begin by defining Human Level Attributes (HLA), which are the characteristics of fictional characters representative of their profile and identity. 
	We collect HLA data from TV Tropes~\cite{tvtropes}, a knowledge-based website dedicated to pop culture, containing information on characters from a variety of sources. Similar to Wikipedia, its content is provided and edited collaboratively by a massive user-base. These attributes are determined by human viewers and their impressions of the characters, and are correlated with human-like characteristics. Furthermore, many tropes include context information (e.g. \textit{jealous girlfriend}) versus high-level personality models such as the Big Five. We believe that TV Tropes is better for our purpose of fictional character modeling than data sources used in works such as \citeauthor{shuster2019engaging} \shortcite{shuster2019engaging} because TV Tropes' content providers are rewarded for correctly providing content through community acknowledgement.
	
	TV Tropes defines \textit{tropes} as attributes of storytelling that the audience recognizes and understands. We use tropes as HLAs to calculate correlations with specific target characters. We collect data from numerous characters from a variety of TV shows, movies, and anime. We filter and keep characters with at least five HLA, as those with fewer are not complex enough to be correctly modeled due to reasons such as lack of data. We end up eliminating 5.86\% of total characters, and end up with 45,821 characters and 12,815 unique HLA, resulting in 945,519 total character-HLA pairs. Each collected character has 20.64 HLAs on average. See Figure~\ref{fig:sheldon_diagram} for an example character and their HLAs.
	
	\subsection{Overall Task}
	Our task is the following, where $t$ denotes ``target":
	
	\smallbreak
	\noindent \textit{Given a target character $c_t$ with HLA set $H_t$, recover the language style of $c_t$ without any dialogue of $c_t$ provided.}
	\smallbreak
	%%JH: need to define what "language style" means? %% also clarify here *why* no data from C_t is provided - why is this necessary? because we use it for testing purposes exclusively.
	%%SF: Defined and discussed below
	
	\noindent For example, if \textit{Sheldon Cooper} from \textit{The Big Bang Theory} is $c_t$, then $H_t$ is the set of HLA on the left side of Figure~\ref{fig:sheldon_diagram}.
	
	We define the language style of a character as its diction, tone, and speech patterns. It is a character specific language model refined from a general language model. We must learn to recover the language style of $c_t$ without its dialogue as our objective is to imitate human-like qualities, and hence the model must understand the language styles of characters based on their traits. If we feed $c_t$'s dialogue during training, the model will likely not effectively learn to imitate language styles based on HLAs, but based on the correlation between text in the training and testing dialogues \cite{joshi2019bert}.
	
\begin{figure}
		\begin{tabular}{c}
			\includegraphics[width=.95\columnwidth]{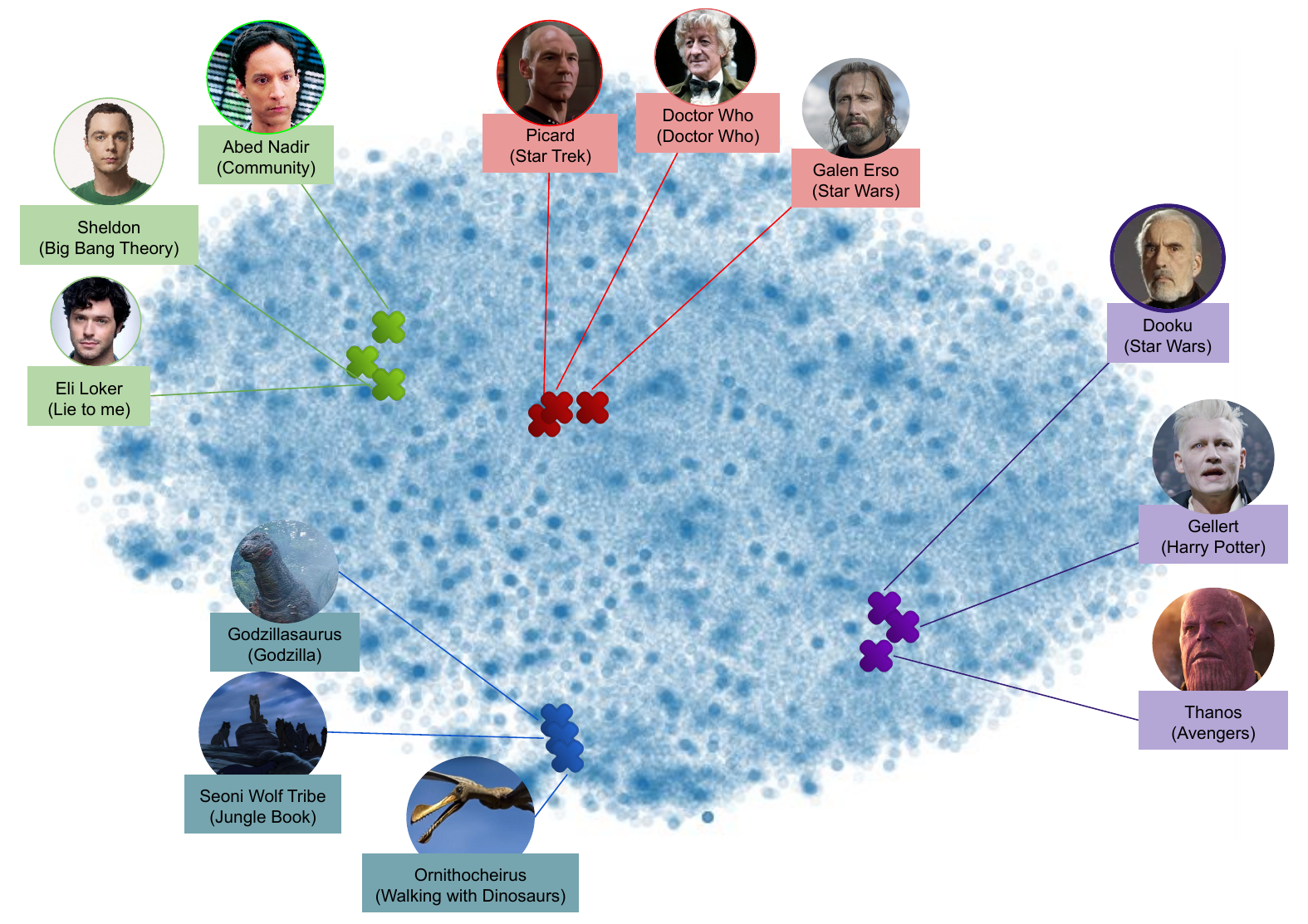}
		\end{tabular}
		\caption{\label{fig:character_space} t-SNE visualization of the character space generated by our Character Space Module (CSM) based on HLAs. %Example characters from four regions are presented.
		}
	\end{figure}

	We define \textit{character space} as the character representations within the HLA latent space (see Figure \ref{fig:character_space}), and the set $C = \{c_1,c_2,...,c_n\}$ as the set of all characters. We define \textit{Observation (OBS)} as the input that is fed into any dialogue model. This can be a single or multiple lines of dialogue along with any additional information. The goal of the dialogue model is to find the best response to this OBS. We show an example of our model's responses imitating five distinct characters in Table~\ref{tab:interaction_example}.
	
	\begin{table}
		\begin{tabular}{c}
			\includegraphics[width=.95\columnwidth]{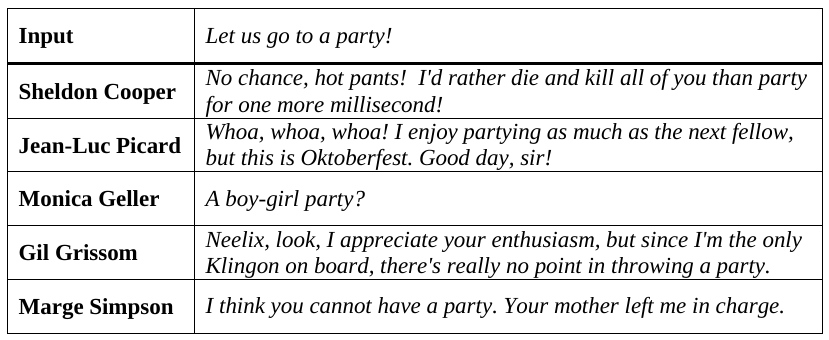}
		\end{tabular}
		\caption{\label{tab:interaction_example} Interaction example using ALOHA-Poly.}
	\end{table}

	\subsection{ALOHA}
\begin{figure}
		\begin{tabular}{c}
			\includegraphics[width=.95\columnwidth]{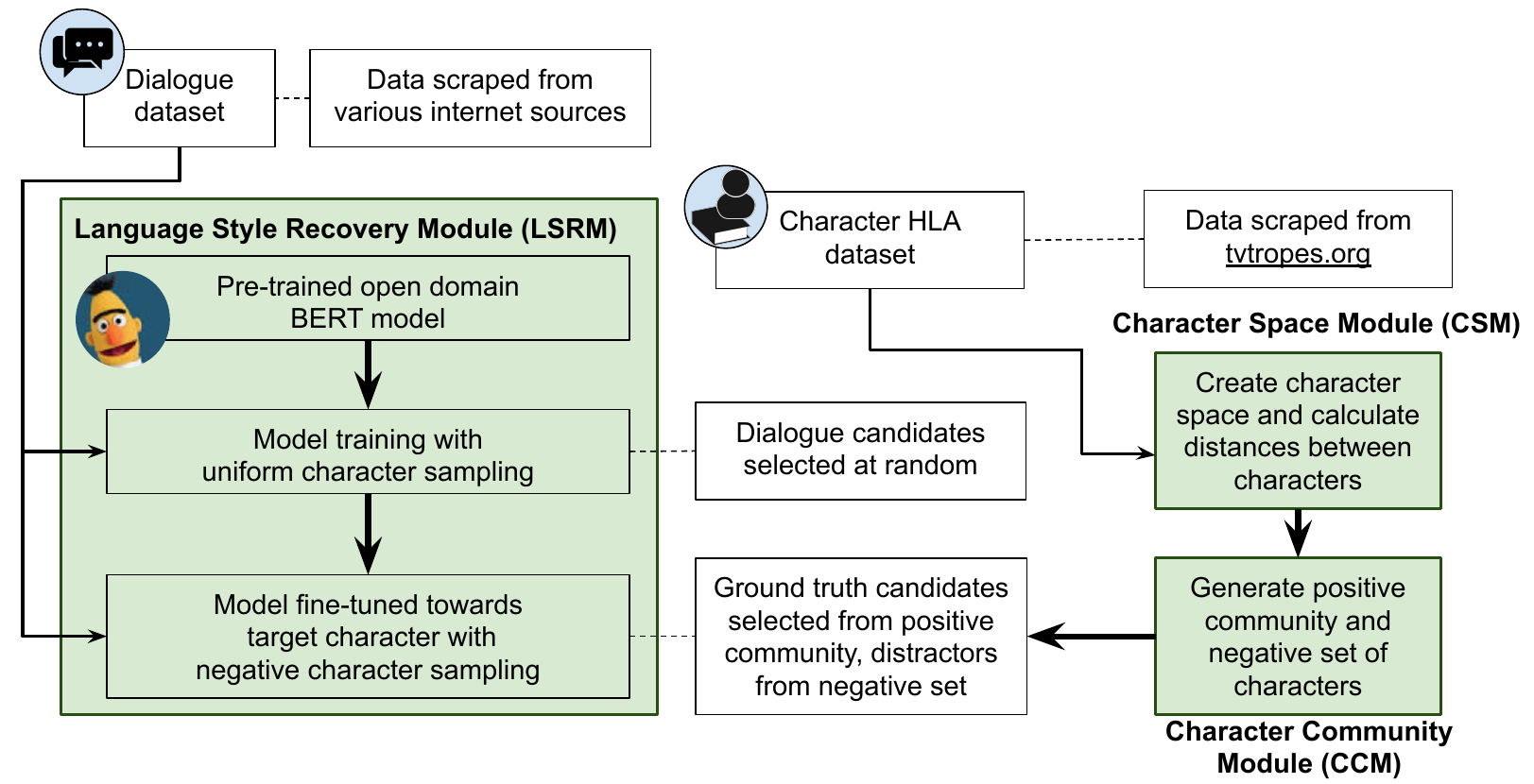} \\
		\end{tabular}
		\caption{\label{fig:system_diagram} Overall system architecture for ALOHA-BERT.}
	\end{figure}

	We propose a three-component system called ALOHA to solve the task (see Figure~\ref{fig:system_diagram}). The first component, Character Space Module (CSM), generates the character space and calculates confidence levels using singular value decomposition \cite{sarwar2000application} between characters $c_j$ (for $j = 1$ to $n$ where $j \neq t$) and $c_t$ in the HLA-oriented neighborhood. 
	
	The second component, Character Community Module (CCM), ranks the similarity between our target character $c_t$ with any other character $c_j$ by the relative distance between them in the character space. 
	%JH2: this is repeated later on so removed here - the notion of positive and negative communities is not even mentioned in the third step here so its unnecessary to describe it here - check
	%%SF2: okay
	%Using these rankings, it identifies \textit{positive} (similar) and \textit{negative} (dissimilar) communities of characters for $c_t$. 
	
	%We define this positive community as the characters that are densely connected internally to the given character within the character space. We treat all other characters outside of this community as belonging to a \textit{negative} set. 
	%These allow us to sample dialogue lines from characters with language styles that are either similar or dissimilar to $c_t$'s language style, respectively.
	
	The third component, Language Style Recovery Module (LSRM), recovers the language style of $c_t$ without its dialogue by training the BERT bi-ranker model \cite{devlin2018bert} and Poly-encoder \cite{humeau2019real} to rank responses from similar characters. This results in two variations of our system, ALOHA-BERT and ALOHA-Poly. Our results demonstrate higher accuracy at retrieving $c_t$'s ground truth response. Our system is able to pick responses which are correct both in context as well as character space.
	
	Hence, the overall process for ALOHA works as follows. First, given a set of characters, determine the character space using the CSM. Next, given a specific target character, determine the positive community and negative set of associated characters using the CCM. Lastly, using the positive community and negative set determined above along with a dialogue dataset, recover the language style of the target. 
	%This is done by retrieving the best response out of a long list of candidate responses for each input dialogue given to ALOHA.

	\subsection{Character Space Module (CSM)}
	CSM learns how to rank characters. We can measure the interdependencies between the HLA variables \cite{hu2008collaborative} and rank the similarity between the TV show characters. We use implicit feedback instead of neighborhood models (e.g. cosine similarity) because it can compute latent factors to transform both characters and HLAs into the same latent space, making them directly comparable. 
	%We call this process \textit{Collaborative Filtering with HLA Measurements}.
	
	%JH2: I don't think this is needed - its implementation detail ?  check
	%%SF2: agreed
	%We set up our character-HLA pairs in matrix form. This process (described below) is completed by modifying the open source project \textit{``python-implicit"} [CITE/LINK]. 
	
	We define a matrix $P$ that contains binary values, with $P_{u,i} = 1$ if character $u$ has HLA $i$ in our dataset, and $P_{u,i} = 0$ otherwise. We define a constant $\alpha$ that measures our confidence in observing various character-HLA pairs as positive. $\alpha$ controls how much the model penalizes the error if the ground truth is $P_{u,i} = 1$. If $P_{u,i} = 1$ and the model guesses incorrectly, we penalize by $\alpha$ times the loss. But if $P_{u,i} = 0$ and the model guesses a value greater than 0, we do not penalize as $\alpha$ has no impact. This is because $P_{u,i} = 0$ can either represent a true negative or be due to a lack of data, and hence is less reliable for penalization. See Equation~\ref{eqn:loss}. We find that using $\alpha=20$ provides decent results.
	%%JH F is not clearly defined - where does this come from and how is it determined? What is F if P is 0, or 1 ... maybe give examples? 
	%%SF: Fixed, see above explanation and F value of 20
	
\begin{figure}
		\begin{tabular}{c}
			\includegraphics[width=.95\columnwidth]{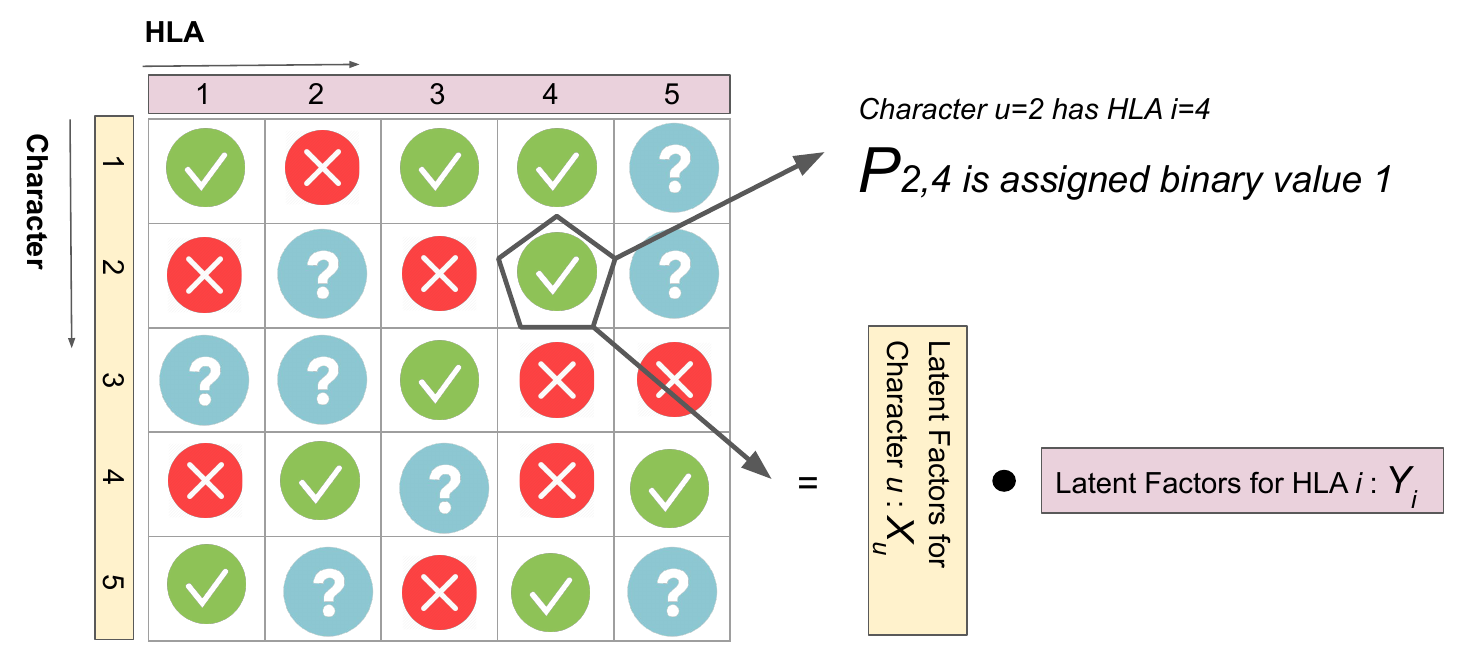} \\
		\end{tabular}
		\caption{\label{fig:matrix_diagram} Illustration of our Collaborative Filtering procedure. Green check-marks indicate a character having an HLA, and `X' indicates otherwise. We randomly mask 30\% of this data for validation, as marked by the `?'. %(described in \textit{Automatic Evaluation}).
		}
	\end{figure}

	We further define two dense vectors $X_u$ and $Y_i$. We call $X_u$ the ``latent factors for character $u$", and $Y_i$ the ``latent factors for HLA $i$". The dot product of these two vectors produces a value ($X_u^TY_i$) that approximates $P_{u,i}$ (see Figure~\ref{fig:matrix_diagram}). This is analogous to factoring the matrix $P$ into two separate matrices, where one contains the latent factors for characters, and the other contains the latent factors for HLAs. We find that $X_u$ and $Y_i$ being 36-dimensional produces decent results. To bring $X_u^TY_i$ as close as possible to $P_{u,i}$, we minimize the following loss function using the Conjugate Gradient Method \cite{takacs2011applications}:
	
	\begin{footnotesize}
		\begin{align}
		loss = {\sum_{u}}{\sum_{i}}({\alpha}P_{u,i} - X_u^TY_i)^2 + \lambda(||X_u||^2 + ||Y_i||^2)
		\label{eqn:loss}
		\end{align}
	\end{footnotesize}
	
	\noindent The first term penalizes differences between the model's prediction ($X_u^TY_i$) and the actual value ($P_{u,i}$). The second term is an L2 regularizer to reduce overfitting. We find $\lambda = 100$ provides decent results for 500 iterations (see Section~\ref{sec:CSM_evaluation}).
	%%JH: need to clarify what algorithm is actually used to do this minimization! 
	%%SF: we use the Conjugate Gradient Method, which I have added above (before the equation)
	
	\subsection{Character Community Module (CCM)}
	CCM aims to divide characters (other than $c_t$) into a \textit{positive} community and a \textit{negative} set. We define this positive community as characters that are densely connected internally to $c_t$ within the character space, and the negative set as the remaining characters. We can then sample dialogue from characters in the negative set to act as the \textit{distractors} (essentially \textit{negative samples}) during LSRM training.
	
	As community finding is an ill-defined problem \cite{fortunato2016community}, we choose to treat CCM as a simple undirected, unweighted graph. We use the values learned in the CSM for $X_u$ and $Y_i$ for various values of $u$ and $i$, which approximate the matrix $P$. Similar to \citeauthor{hu2008collaborative} \shortcite{hu2008collaborative}, we can calculate the correlation between two rows (and hence two characters).
	
\begin{figure}
		\begin{tabular}{ccc}
			\multicolumn{1}{c}{\includegraphics[width=.95\columnwidth]{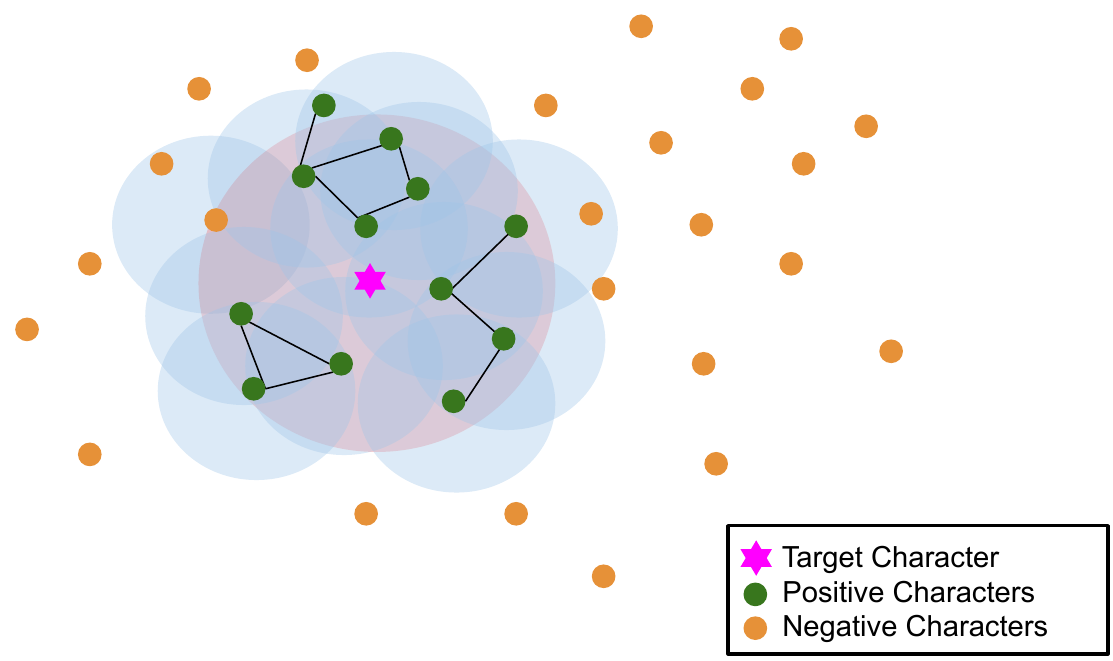}}
		\end{tabular}
		\caption{\label{fig:cluster_diagram} Illustration of the two-level connection representation procedure, using a minimum frequency of two. The transparent red circle indicates the first level set ($S^{FL}$), while the blue ones indicate the sets $S^{SL}_i$. The lines indicate connections between the characters within the community structure of $c_t$.}
	\end{figure}

	We then employ a two-level connection representation by ranking all characters against each other in terms of their correlation with $c_t$. For the first level, the set $S^{FL}$ is the top 10\% most highly correlated characters with $c_t$ out of the 45,820 total other characters we have HLA data for. For the second level, for each character $s_i$ in $S^{FL}$, we determine the 30 most heavily correlated characters with $s_i$ as set $S^{SL}_i$. The positive set $S^{pos}$ are the characters present in at least 10 $S^{SL}_i$ sets. We call this value 10 the \textit{minimum frequency}. All other characters in our dialogue dataset make up the negative set $S^{neg}$. These act as our \textit{positive} community and \textit{negative} set, respectively. See Figure~\ref{fig:cluster_diagram} for an example.
	
	%%JH I suggest removing algorithm 1 - its not really necessary
	%%SF: Maybe we can move it to the appendix. I thought this was necessary as the entire CCM process is somewhat difficult to understand, and this algorithm makes it easily reproducible (e.g. think back to the Yao WordNet algorithm that made it very easy to reimplement for the SMERTI paper).
	%\begin{figure}
	%\begin{tabular}{ccc}
	%\multicolumn{1}{c}{\includegraphics[width=.95\columnwidth]{algorithm1.png}}
	%\end{tabular}
	%\end{figure}
	
	\subsection{Language Style Recovery Module (LSRM)\label{sec:LSRM}}
	LSRM creates a dialogue agent that aligns with observed characteristics of human characters by using the positive character community and negative set determined in the CCM, along with a dialogue dataset, to recover the language style of $c_t$ without its dialogue. We use the BERT bi-ranker model from the Facebook ParlAI framework \cite{miller2017parlai} and the Poly-encoder \cite{humeau2019real}, where the models have the ability to retrieve the best response out of 20 candidate responses. These are trained to produce LSRM-BERT and LSRM-Poly, respectively. \cite{dinan2019second,urbanek2019learning,zhang2018personalizing} choose 20 candidate responses, and for comparison purposes, we do the same.
	
	\subsubsection{BERT} \cite{devlin2018bert} is first trained on massive amounts of unlabeled text data. It jointly conditions on text on both the left and right, which provides a deep bi-directional representation of sentence inference. BERT is proven to perform well on a wide range of tasks by simply fine-tuning on one additional layer. We are interested in its ability to predict the next sentence, called \textit{Next Sentence Prediction}. We perform further fine-tuning on BERT for our target character language style retrieval task to produce our LSRM-BERT model by optimizing both the encoding layers and the additional layer. We use BERT to create vector representations for the OBS and for each candidate response. By passing the first output of BERT's 12 layers through an additional linear layer, these representations can be obtained as 768-dimensional sentence-level embeddings. It uses the dot product between these embeddings to score candidate responses and is trained using the ranking loss.
	%%JH: need a bit more detail on how BERT works - its not enough to cite the paper - you need to give at least a high level overview of what its actually composed of. I am not clear what the "768-dimensional embeddings" mean - this needs more clarification
	%%SF: see above for more detailed explanation
	
	\subsubsection{Poly-encoder} \cite{humeau2019real} is a transformer architecture that learns global rather than local level token features to perform attention on. The model has state-of-the-art accuracy on response retrieval on the Persona-Chat dataset. As in the Bi-encoder, a given candidate response is first encoded into a vector. Then, softmax attention against multiple context vectors encoded from the input observation is performed to compute the final score.
	
	\subsubsection{Candidate response selection} is similar to the procedure from previous work done on grounded dialogue agents \cite{zhang2018personalizing,urbanek2019learning}. Along with the ground truth response, we randomly sample 19 \textit{distractor} responses from other characters from a uniform distribution of characters, and call this process \textit{uniform character sampling}. Based on our observations, this random sampling provides multiple context correct responses. Hence, the BERT bi-ranker model is trained by learning to choose context correct responses, and the model learns to recover a domain-general language model that includes training on every character. This results in a \textit{Uniform Model} that can select context correct responses, but not responses corresponding to a target character with specific HLAs.
	
	We then fine-tune on the above model to produce our LSRM-BERT model with a modification: we randomly sample the 19 \textit{distractor} responses from only the negative character set instead. We choose the responses that have similar grammatical structures and semantics to the ground truth response, and call this process \textit{negative character sampling}. This guides the model away from the language style of these negative characters to improve performance at retrieving responses for target characters with specific HLAs.
	
	We train a second version of our LSRM, \textit{LSRM-Poly}, by training Poly-encoder directly on HLA-Chat using negative character sampling following the same procedure as above with 19 distractor responses. Our results demonstrate higher accuracy from both ALOHA-BERT and ALOHA-Poly variations of our system at retrieving the correct response from character $c_t$, which is the ground truth.
	
	\section{Experiment}
	\subsection{Dialogue Dataset and HLA-Chat\label{sec:dialogue_data}}
	To train the Uniform Model and LSRM, we collect dialogues from 327 major characters (a subset of the 45,821 characters we have HLA data for) in 38 TV shows from various existing sources of clean data on the internet, resulting in a total of 1,042,647 dialogue lines.
	%%JH: is the 327 a subset of the 45,821 above? need to clarify this 
	%%SF: Yes, confirmed with Aaron and clarified above
	We use a setup similar to the Persona-Chat dataset \cite{zhang2018personalizing} and Cornell Movie-Dialogs Corpus \cite{danescu2011chameleons}, as our collected dialogues are also paired in terms of valid conversations.\footnote{Our dataset has much more dialogue per character compared to Persona-Chat and Cornell Movie-Dialogs Corpus as we need sufficient data to learn each character's language style.} See the right side of Figure~\ref{fig:sheldon_diagram} for an example of these dialogue lines. We combine these dialogue lines with our collected HLA (tropes) data for these characters to form our proposed dataset, HLA-Chat. 
	%%JH2: not needed? check 
	%%SF2: agreed, this is stated in the introduction
	%The modification is that while the persona-chat dialogues are traceable to persona profiles, our dialogues are traceable to specific characters with many HLAs as descriptors. Rather than associating each dialogue with a single persona or HLA, we associate it with a specific character that exhibits a combination of various HLAs.
	
	\subsection{HLA Observation Guidance (HLA-OG)\label{sec:HLA-OG}}
	We define \textit{HLA Observation Guidance} (HLA-OG) as explicitly passing a small subset of the most important HLAs of a given character as part of the OBS rather than just an initial line of dialogue. This is adapted from the process used in \citeauthor{zhang2018personalizing} \shortcite{zhang2018personalizing} and \citeauthor{wolf2019transfertransfo} \shortcite{wolf2019transfertransfo} which we call \textit{Persona Profiling}. Specifically, we pass eight HLAs that are randomly drawn from the top 40 most important HLAs of the character. We train the Uniform Model using \textit{No HLA-OG} by explicitly passing eight HLAs of \textit{`none'} along with the initial line of dialogue as the OBS. We use HLA-OG during training of the LSRM (both BERT and Poly-encoder variations) and testing of all models. This is because the baselines (see Section~\ref{sec:baselines}) already follow a similar process (\textit{Persona Profiling}) for training. For testing, HLA-OG is necessary as it provides information about which HLAs the models should attempt to imitate in their response selection. Just passing an initial line of dialogue without HLAs replicates a typical dialogue response task based only on context correctness. See Table~\ref{tab:HLA-OG_table}. 
	%JH2 superfluous 
	%By comparing the results of this with the results of the LSRM with HLA-OG, we can determine whether explicitly giving the LSRM knowledge of the HLAs during observation in training improves its performance.
	
	\begin{table}
		\begin{tabular}{ccc}
			\multicolumn{1}{c}{\includegraphics[width=.95\columnwidth]{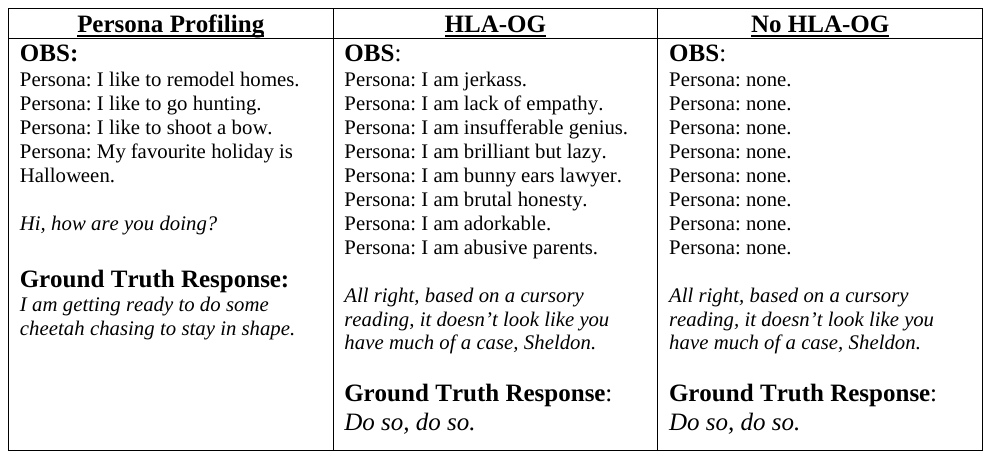}}\\
		\end{tabular}
		\caption{\label{tab:HLA-OG_table} Example for Persona Profiling, HLA-OG, and No HLA-OG. All lines under OBS are fed together as input to the language style retrieval model.}
	\end{table}
	
	\subsection{Training Details\label{sec:training_details}}
	\subsubsection{BERT bi-ranker} is a baseline model trained by us on the Persona-Chat dataset. Similar to \citeauthor{zhang2018personalizing} \shortcite{zhang2018personalizing}, we cap the length of the OBS at 360 tokens and the length of each candidate response at 72 tokens.\footnote{\textit{Tokens} here refer to the WordPiece tokens used by BERT.} We use a batch size of 80, learning rate of 5e-5, and perform warm-up updates for 1000 iterations. The learning rate scheduler uses SGD optimizer with Nesterov's accelerated gradient descent \cite{sutskever2013importance} and is set to have a decay of 0.4 and to reduce on plateau.\footnote{We are able to recover up to 78\% Hits@1 accuracy on Persona-Chat (see Section~\ref{sec:key_metrics}).} We initialize using pretrained fastText \cite{fasttext} embeddings.
	
	\subsubsection{Uniform Model} is trained using BERT's pretrained weights on the dialogue data discussed in Section~\ref{sec:dialogue_data} using uniform character sampling. We use the same hyperparameters as the BERT bi-ranker along with half-precision operations (i.e. float16 operations) to increase batch size as recommended \cite{humeau2019real}. We initialize using pretrained fastText embeddings.
	
	\subsubsection{LSRM-BERT} is produced by finetuning on the Uniform Model discussed above using HLA-Chat and negative character sampling. We use the same hyperparameters as the BERT bi-ranker with the same half-precision operations as above. We refer to our BERT model as \textit{ALOHA-BERT}.
	
	\subsubsection{LSRM-Poly} is produced by training the Poly-encoder directly on HLA-Chat using negative character sampling. \citeauthor{humeau2019real} \shortcite{humeau2019real} introduced an effective pretraining procedure on Reddit data which we fine-tune on. Other than using a smaller batch size of 80, we adapt all parameters used in \citeauthor{humeau2019real} \shortcite{humeau2019real}: Adam optimizer with learning rate of 2e-4, $\beta1 = 0.9$, $\beta2 = 0.98$, no L2 weight decay, linear learning rate warmup, and inverse square root decay of the learning rate. We refer to our Poly-encoder model as \textit{ALOHA-Poly}.
	
	\section{Evaluation}
	\subsection{CSM Evaluation\label{sec:CSM_evaluation}}
	We begin by evaluating the ability of the CSM component of our system to correctly generate the character space. To do so, during training, 30\% of the character-HLA pairs (which are either 0 or 1) are masked, and this is used as a validation set (see Figure~\ref{fig:matrix_diagram}). For each character $c$, the model generates a list of the 12,815 unique HLAs ranked similarly to \citeauthor{hu2008collaborative} \shortcite{hu2008collaborative} for $c$. We look at the recall of our CSM model, which measures the percentage of total ground truth HLAs (over all characters $c$) present within the top N ranked HLAs for all $c$ by our model. That is:
	
	\begin{footnotesize}
		\begin{align}
		recall = \frac{{\sum_{c}}|HLA_{c}^{gt} \cap HLA_{c}^{tN}|}{{\sum_{c}}|HLA_{c}^{gt}|}
		\label{eqn:recall}
		\end{align}
	\end{footnotesize}
	
	\noindent where $HLA_{c}^{gt}$ are the ground truth HLAs for $c$, and $HLA_{c}^{tN}$ are the top N ranked HLAs by the model for $c$. We use $N = 100$, and our model achieves 25.08\% recall.
	
	To inspect the CSM performance, we use the T-distributed Stochastic Neighbor Embedding (t-SNE) \cite{maaten2008visualizing} to reduce each high-dimensionality data point to two-dimensions via Kullback-Leibler Divergence \cite{kullback1951information}. This allows us to map our character space into two-dimensions, where similar characters from our embedding space have higher probability of being mapped close by. We sampled characters from four different groups or regions. As seen in Figure~\ref{fig:character_space}, our learned character space effectively groups these characters, as similar characters are adjacent to one another in four regions.
	
	\subsection{Automatic Evaluation Setup\label{sec:auto_eval_setup}}
	
	\subsubsection{Five-Fold Cross Validation} is used for the training and testing of the Uniform Model and two LSRM variations. The folds are divided randomly by the TV shows in our dialogue data. We use the dialogue data for 80\% of these shows as the four-folds for training, and the dialogue data for the remaining 20\% as the fifth-fold for testing. The dialogue data used is discussed in Section~\ref{sec:dialogue_data}. This ensures no matter how our data is distributed, each part of it is tested, allowing our evaluation to be more robust to different characters. See Table~\ref{tab:five_fold} for detailed statistics.
	
	\begin{table}
		\begin{tabular}{ccc}
			\multicolumn{1}{c}{\includegraphics[width=.95\columnwidth]{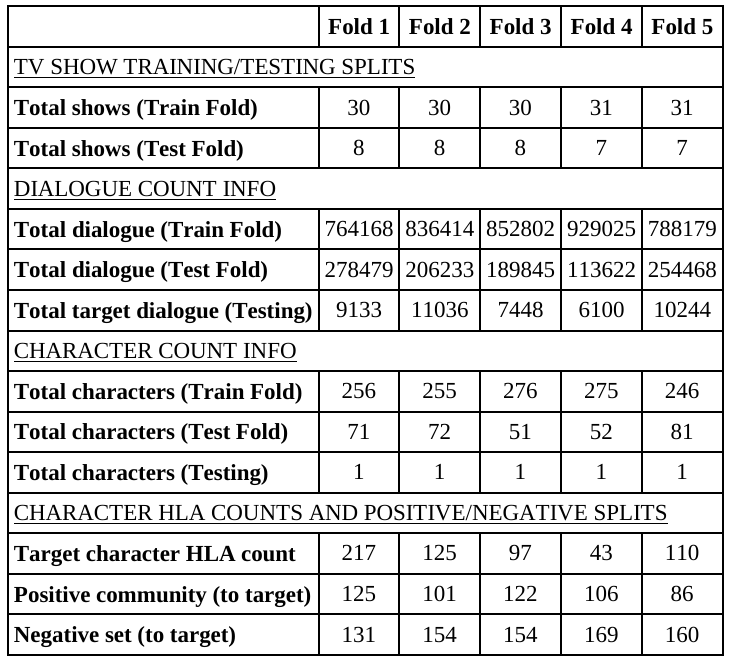}}\\
		\end{tabular}
		\caption{\label{tab:five_fold} Detailed five-fold cross validation statistics.}
	\end{table}
	
	\subsubsection{Five Evaluation Characters} are chosen, one from each of the five testing sets described above. Each is a well-known character from a separate TV show, and acts as a target character $c_t$ for evaluation of every model. We choose \textit{Sheldon Cooper} from \textit{The Big Bang Theory}, \textit{Jean-Luc Picard} from \textit{Star Trek}, \textit{Monica Geller} from \textit{Friends}, \textit{Gil Grissom} from \textit{CSI}, and \textit{Marge Simpson} from \textit{The Simpsons}. We choose characters of significantly different identities and profiles (intelligent scientist, ship captain, outgoing friend, police leader, and responsible mother, respectively) from shows of a variety of genres to ensure that we can successfully recover the language styles of various types of characters. We choose well-known characters because humans require knowledge on the characters they are evaluating (see Section~\ref{sec:human_eval}).
	
	For each of these five evaluation characters, all the dialogue lines from the character act as the ground truth responses. The initial dialogue lines are the corresponding dialogue lines to which these ground truth responses are responding. For each initial dialogue line, we randomly sample 19 other candidate responses from the associated testing set using uniform character sampling. Note that this is for evaluation, and hence we use the same uniform character sampling method for all models including ALOHA. The use of negative character sampling is only in ALOHA's training.

	\subsection{Baselines\label{sec:baselines}}
	We compare against four dialogue system baselines: Kvmemnn, Feed Yourself, Poly-encoder, and a BERT bi-ranker baseline trained on the Persona-Chat dataset using the same training hyperparameters (including learning rate scheduler and length capping settings) described in Section~\ref{sec:training_details}.\footnote{See Section~\ref{sec:related_works} for more details about the first three models.} For the first three models, we use the provided pretrained (on Persona-Chat) models. For the Poly-encoder baseline, we use the official model trained on ConvAI2. We evaluate all four baselines on our five evaluation characters discussed in Section~\ref{sec:auto_eval_setup}.
	
	\subsection{Key Evaluation Metrics\label{sec:key_metrics}}
	\subsubsection{Hits@n/N} is the accuracy of the correct ground truth response being within the top $n$ ranked candidate responses out of $N$ total candidates. We measure Hits@1/20, Hits@5/20, and Hits@10/20.
	
	\subsubsection{Mean Rank} is the average rank that a model assigns the ground truth response among the 20 total candidates.
	
	\subsubsection{Mean Reciprocal Rank (MRR)} looks at the mean of the multiplicative inverses of the rank of each correct answer out of a sample of queries $Q$:
	
	\begin{footnotesize}
		\begin{align}
		MRR = \frac{1}{|Q|}{\sum_{i=1}^{|Q|}}\frac{1}{rank_i}
		\label{eqn:MRR}
		\end{align}
	\end{footnotesize}
	
	\noindent where $rank_i$ refers to the rank position of the correct response for the $i$-th query, and $|Q|$ refers to the total number of queries in $Q$.
	
	\subsubsection{\textbf{$F_1$}-score} equals $2 * \frac{precision*recall}{precision+recall}$. For dialogue, precision is the fraction of words in the chosen response contained in the ground truth, and recall is the fraction of words in the ground truth response contained in the chosen one.
	
	\subsubsection{BLEU} \cite{papineni2002} generally indicates how close two pieces of text are in content and structure, with higher values indicating greater similarity. We report our final BLEU scores as the average scores of 1 to 4-grams.
	
	%\subsubsection{Perplexity} is a measure of text fluency and is computed as %$\frac{1}{m}{\sum_{i=1}^{m}}log[p(w_i)]$, where $w_1$ to $w_m$ are the individual words of %a sentence \textbf{w}. Lower perplexity values indicate greater fluency.
	
	%%JH: need to also describe the automatic evaluation set-up
	%%SF: I believe this is the exact same as the human evaluation setup, except the BERT model ranks all 20 candidate responses rather than just choosing the best. I have renamed the previous "Five-Fold Cross Validation" section to "Automatic Evaluation Setup", and added explanation for this there
	%%JH: also - solve the baseline problem - I think its really sketchy to throw out the two outliers just because they scored 0 - was there anything else wrong with these two? in any case, you have to state this and explain what happened.
	%%SF: We decided to add the two outliers back in, as the only reason we excluded them was because they scored 0 - hence, addressed
	\subsection{Human Evaluation Setup\label{sec:human_eval}}
	We conduct a human evaluation to get an upper bound on expected performance with 12 participants, 8 male and 4 female, who are affiliated project researchers aged 20-39 at the University of Waterloo. We choose the same five evaluation characters as in Section~\ref{sec:auto_eval_setup}. To control bias, each participant evaluates one or two characters. For each character, we randomly select 10 testing samples (each includes an initial line of dialogue along with 20 candidate responses, one of which is the ground truth) from the same testing data for the automatic evaluation discussed in Section~\ref{sec:auto_eval_setup}. 
	
\begin{figure}
		\begin{tabular}{c}
			\includegraphics[width=.95\columnwidth]{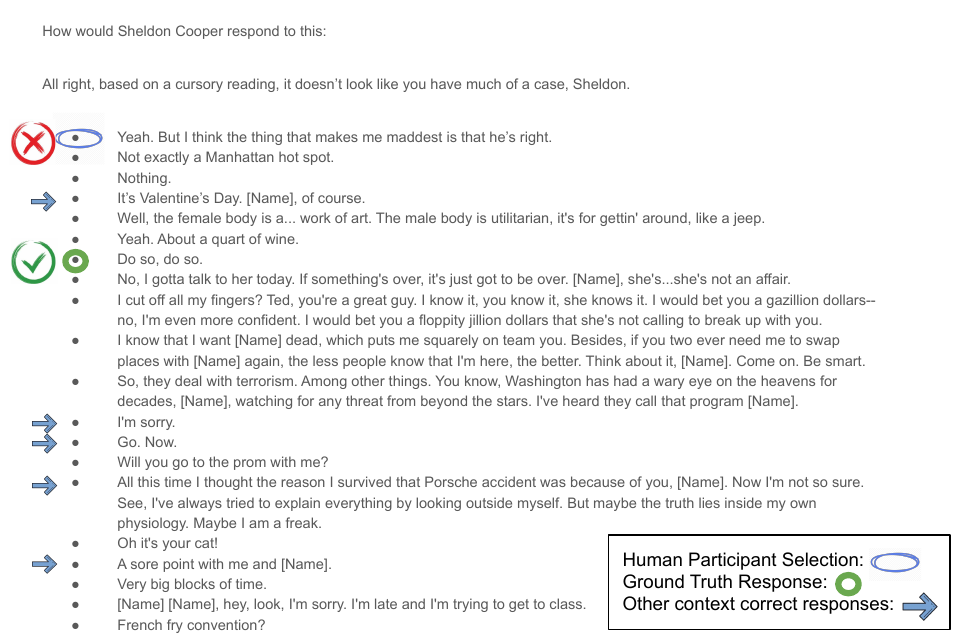}
		\end{tabular}
		\caption{\label{fig:human_eval_example} Example of what a human participant sees on each page of the questionnaire, along with their chosen response and the ground truth. As seen, there are multiple context correct (but not necessarily HLA correct) candidate responses. In this case, Sheldon as a character should not be able to admit his mistake.}
	\end{figure}

	These ten samples make up a single questionnaire presented in full to each participant evaluating the corresponding character, and the participant is asked to select the single top response they think the character would most likely respond with for each of the ten initial dialogue lines. See Figure~\ref{fig:human_eval_example} for an example. We mask any character names within the candidate responses to prevent human participants from using names to identify which show the response is from.
	
	%%JH2: I really like this figure but it might need to be cut and moved to appendix
	%%SF2: will do so if we still cannot get under the page limit
	
	Each candidate is prescreened to ensure they have sufficient knowledge of the character to be a participant. We ask three prescreening questions where the participant has to identify an image, relationship, and occupation of the character. All 12 of our participants passed the the prescreening. %%JH: the following could be made much shorter here, with the full questions in the appendix. You can say "We ask three prescreening questions in which the participant had to identify an image, relationship and occupation of the character". 
	%%JH: also need to say that all 12 participants passed the screening tests
	%%SF: addressed both points
	
	%\begin{enumerate}
	%\item{\textit{``Select which one is $c_i$"}, where the candidate must choose which image is $c_i$ out of multiple images of different characters.}
	%\item{\textit{``Which one best describes $c_i$'s relationship to $c_j$?}, where $c_j$ is another major character in the same show, and the candidate must choose from a list of relationships (e.g. ``Commander", ``Brother", ``Friend", and so forth).}
	%\item{\textit{``Which one best describes $c_i$'s occupation most of the time?"}, where the candidate must choose from a list of occupations related to show's context. (e.g., ``Science Officer`` and ``Ship Captain`` in sci-fi context).}
	%\end{enumerate}
	
	\section{Results and Analysis}
	
	\subsection{Evaluation Results}
	%%JH: need to have a table showing the amount of data being used here - as in the SMERTI paper
	%%SF: NEED DATA FROM AARON (will try to get it ASAP)
	%%JH: need to describe the dialogue data - how much, etc. Also mentioned in "Five Fold ..." section - not sure where it fits best
	%%SF: As stated earlier, this is addressed earlier in the HLA section (which I have now moved into a "Dialogue Data" section). However, I agree that we should give the total number of dialogue lines (either exact or approximate) (ADDED in "dialogue data" section)
	\begin{table}
		\begin{tabular}{ccc}
			\multicolumn{1}{c}{\includegraphics[width=.95\columnwidth]{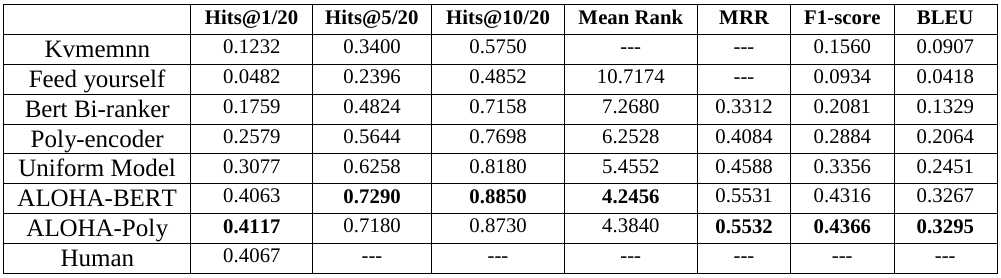}}\\
		\end{tabular}
		\caption{\label{tab:results_1} Average automatic evaluation on HLA-OG and human evaluation results. Bold indicates the best performance.}
	\end{table}
	
	\begin{table}
		\begin{tabular}{ccc}
			\multicolumn{1}{c}{\includegraphics[width=.95\columnwidth]{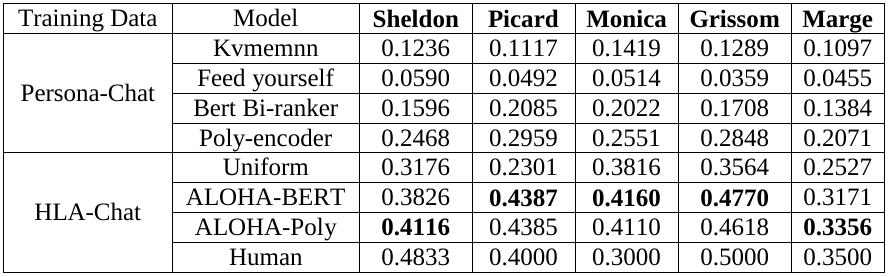}}\\
		\end{tabular}
		\caption{\label{tab:results_2} Average Hits@1/20 scores by evaluation character on HLA-OG data. Bold indicates the best performance (excluding humans).}
	\end{table}
	
	%%JH2: probably don't need this here - move to appendix
	%%SF2: done
	\commentout{
		\begin{figure}
			\begin{tabular}{ccc}
				\multicolumn{1}{c}{\includegraphics[width=.95\columnwidth]{relative_frequency_histogram.png}}\\
			\end{tabular}
			\caption{\label{fig:relative_freq} Relative frequency histogram of correctly chosen responses for the human evaluation.}
		\end{figure}
	}
	
	Table~\ref{tab:results_1} shows average results of our automatic and human evaluations. Table~\ref{tab:results_2} shows average Hits@1/20 scores by evaluation character. See Table~\ref{tab:interaction_example} for a demo interaction example between a human and ALOHA-Poly for all five evaluation characters.
	
	\subsection{Evaluation Challenges}
	The evaluation of our task (retrieving the language style of a specific character) is challenging and hence the five-fold cross validation is necessary for the following reasons:
	
	1. The ability to choose a context correct response without attributes of specific characters may be hard to separate from our target metric, which is the ability to retrieve the correct response of a target character by its HLAs. However, from manual observation, we noticed that in the 20 chosen candidate responses, there are typically numerous context correct responses, but only one ground truth for the target character (for an example, see Figure~\ref{fig:human_eval_example}).
	
	To investigate this, we randomly chose 50 sets of input and candidate responses (a total of 1000 candidate responses: 10 sets per target character and 20 responses per set), and manually labelled the number of context correct responses for each set. We found a total of 333 context correct responses (79, 71, 68, 53, 62 for characters 1 to 5 respectively) which means an average of 6.66 (out of 20) per input, and so a random guess over these context-correct responses would give an accuracy of 15\%. Our empirical results indicate human accuracy is around 40\%, demonstrating that humans make a choice relying on much more than just context correctness. Both ALOHA variations perform similarly (around 41\%) and show that human performance is seemingly achievable by our system.
	
	2. Retrieving responses for the target character depends on the other candidate responses. For example, dialogue retrieval performance for Grissom from CSI, which is a crime/police context, is higher than other evaluation characters (see Table~\ref{tab:results_2}), potentially due to other candidate responses not falling within the same crime/police context.
	
	\subsection{Performance: ALOHA vs. Humans}
	As observed from Tables~\ref{tab:results_1} and~\ref{tab:results_2}, ALOHA (both variations) has a performance relatively close to humans. Human Hits@1/20 scores have a mean of 40.67\% and a median over characters of 40\%. The limited human evaluation size limits what can be inferred, but it indicates the problem is solved to the extent that ALOHA is able to slightly outperform humans on two folds and perform closely on another two folds. Even humans do not perform extremely well, demonstrating this task is more difficult than typical dialogue retrieval tasks \cite{urbanek2019learning,dinan2019second}.
	
	Looking more closely at each character from Table~\ref{tab:results_2}, we can see that human evaluation scores are higher for Sheldon and Grissom. This may be due to these characters having more distinct personalities, making them more memorable. ALOHA performs worse on Sheldon compared to humans. This is possibly due to the large number of Sheldon's HLAs (217) compared to the other four evaluation characters (average of 93.75), along with the limited amount of HLAs we are using for guidance due to the models' limited memory.
	
	We also look at Pearson correlation values of the Hits@1/20 scores across the five evaluation characters. For human versus Uniform Model, this is 0.047 (heavily uncorrelated), demonstrating that the Uniform Model, without knowledge of HLAs, fails to imitate human impressions. For human versus ALOHA-BERT and ALOHA-Poly, these are 0.4149 and 0.5468, respectively, demonstrating that ALOHA is able to retrieve character responses somewhat similarly to human impressions. The difference between the ALOHA variations and the Uniform Model is based on the additional knowledge of the HLAs (e.g. by using HLA-OG and negative instead of uniform character sampling). This demonstrates that HLAs are indeed an accurate method of modeling human impressions of character attributes and that ALOHA is able to effectively use them to improve upon dialogue retrieval performance.
	
	\subsection{Performance: ALOHA vs. Baselines}
	ALOHA, combined with HLA-Chat, achieves a significant improvement on the target character language style retrieval task compared to the baseline open-domain chatbot models across all five folds. As observed from Tables~\ref{tab:results_1} and~\ref{tab:results_2}, ALOHA achieves a significant boost in Hits@n/N accuracy and other metrics for retrieving the correct responses from five diverse characters (see Section~\ref{sec:auto_eval_setup}). Paired t-tests between the Hits@1/20 scores of ALOHA-BERT against BERT Bi-ranker and ALOHA-Poly against Poly-encoder across all five evaluation folds are statistically significant with p-values of 0.0004 and less than 0.0001, respectively, showing that the consistent improvement is meaningful.
	
	\subsection{Performance: ALOHA vs. Uniform Model}
	We observe a noticeable improvement in performance between ALOHA and the Uniform Model in recovering the language styles of specific characters that is consistent across all five folds (see Tables~\ref{tab:results_1} and \ref{tab:results_2}). Paired t-tests between the Hits@1/20 scores of ALOHA-BERT and ALOHA-Poly against the uniform model across all five evaluation folds are statistically significant with p-values of 0.0329 and 0.0234, respectively, showing that the consistent improvement is meaningful. This indicates that lack of knowledge of HLAs limits the ability of the model to successfully recover the language style of specific characters. We claim that, to the best of our knowledge, we have made the first step in using HLA-based character dialogue clustering to improve upon personality learning for chatbots.
	
	ALOHA demonstrates an accuracy boost for all five evaluation characters, showing that the system is robust and stable and has the ability to recover the dialogue styles of fictional characters regardless of the character's profile and identity, genre of the show, and context of the dialogue.
	
	\section{Conclusion and Future Work}
	We proposed Human Level Attributes (HLAs) as a novel approach to model human-like attributes of characters, and collected a large volume of dialogue data for various characters with complete HLA profiles which we release in a dataset, HLA-Chat. We also proposed and evaluated a system, ALOHA, that uses HLAs to recommend tailored responses by specific characters. We demonstrated both ALOHA-BERT and ALOHA-Poly's outperformance of the baselines, and their ability to effectively recover language styles of various characters, showing promise for learning character or personality styles. Further, we demonstrated ALOHA's slight outperformance of humans for two out of five evaluation characters, with close performance on two others. ALOHA was shown to be stable regardless of the character's identity, genre of show, and context of dialogue, and ALOHA-Poly was shown to be particularly robust.
	
	Potential directions for future work include training ALOHA with a \textit{multi-turn response} approach \cite{zhang2018personalizing} that tracks dialogue over multiple responses. Another potential is the modeling of the dialogue counterpart (e.g. the dialogue of other characters speaking to the target character). Further, performing \textit{semantic text exchange} on the chosen response with a model such as SMERTI \cite{feng-etal-2019-keep} may improve the ability of ALOHA to converse with humans. This is because the response may be context and HLA correct, but incorrect semantically (e.g. it may say the weather is \textit{sunny} when it is actually \textit{rainy}). HLA-aligned generative models is another area of exploration. Typically, generative models produce text that is less fluent, but further work in this area may lead to better results. Lastly, a more diverse and larger participant pool is required due to the limited size of our human evaluation. We can also investigate other factors affecting human performance on specific characters such as their familiarity with TV series (e.g. they may be more familiar with more recent shows).
	
	%\noindent\textbf{Acknowledgments}
	\section*{Acknowledgments}
	We thank our anonymous reviewers, study participants, and Huawei Technologies Co., Ltd. for financial support.
	
	\fontsize{9.0pt}{10.0pt} \selectfont
	\bibliography{AAAI-LiA.9168}
	\bibliographystyle{aaai}
	
\end{document}